\documentclass{article}

\usepackage[final]{nips_nonotice}

\usepackage[utf8]{inputenc} 
\usepackage[T1]{fontenc}    
\usepackage{hyperref}       
\usepackage{url}            
\usepackage{booktabs}       
\usepackage{nicefrac}       
\usepackage{microtype}      
\usepackage{amsfonts}
\usepackage{amsmath}
\usepackage{amssymb}
\usepackage{wrapfig}
\usepackage{mathtools}
\usepackage{subcaption}
\usepackage[inline]{enumitem}   
\usepackage[skip=0pt]{caption}  

\usepackage[disable]{todonotes}

\newif\ifanony  
\anonyfalse

\usepackage{natbib}
\bibpunct{(}{)}{;}{a}{,}{,}

\AtBeginDocument{

}

\def\O{\mathcal{O}} 
\def\A{\mathcal{A}} 
\def\rest{\hat{r}}  
\newcommand{\abs}[1]{\left|#1\right|}
\DeclarePairedDelimiter{\bdelims}{[}{]}

\DeclarePairedDelimiter{\pdelims}{(}{)}
\newcommand{\of}[1]{\pdelims*{#1}}
\newcommand{\db}{\mathcal{D}}

\def\plotsize{0.96}  

\title{Deep Reinforcement Learning \\ from Human Preferences}
\author{
  Paul F Christiano \\
  OpenAI \\
  \texttt{paul@openai.com} \\
  \And
  Jan Leike \\
  DeepMind \\
  \texttt{leike@google.com} \\
  \And
  Tom B Brown \\
  \texttt{nottombrown@gmail.com} \\
  \And
  Miljan Martic \\
  DeepMind \\
  \texttt{miljanm@google.com} \\
  \And
  Shane Legg \\
  DeepMind \\
  \texttt{legg@google.com} \\
  \And
  Dario Amodei \\
  OpenAI \\
  \texttt{damodei@openai.com} \\
}
\date{\today}

\begin{document}

\maketitle

\begin{abstract}%
For sophisticated reinforcement learning~(RL) systems to interact usefully with real-world environments,
we need to communicate complex goals to these systems.
In this work, we explore goals defined in terms of (non-expert) human preferences between pairs of trajectory segments.
We show that this approach can effectively solve complex RL tasks without access to the reward function,
including Atari games and simulated robot locomotion,
while providing feedback on less than 1\% of our agent's interactions with the environment.
This reduces the cost of human oversight far enough that it can be practically applied to state-of-the-art RL systems.
To demonstrate the flexibility of our approach,
we show that we can successfully train complex novel behaviors
with about an hour of human time.
These behaviors and environments are considerably more complex
than any which have been previously learned from human feedback.
\end{abstract}

\section{Introduction}
\label{sec:introduction}

Recent success in scaling reinforcement learning~(RL) to large problems
has been driven in domains that have a well-specified reward function~\citep{Mnih15,Mnih16,Silver16}.
Unfortunately, many tasks involve goals that are complex, poorly-defined, or hard to specify.
Overcoming this limitation would greatly expand the possible impact of deep RL
and could increase the reach of machine learning more broadly.

For example, suppose that we wanted to use reinforcement learning
to train a robot to clean a table or scramble an egg.
It's not clear how to construct a suitable reward function,
which will need to be a function of the robot's sensors.
We could try to design a simple reward function that approximately captures the intended behavior,
but this will often result in behavior that optimizes our reward function without actually
satisfying our preferences.
This difficulty underlies recent concerns about misalignment
between our values and the objectives of our RL systems~\citep{Bostrom14,Russell16,Amodei16}.
If we could successfully communicate our actual objectives to our agents,
it would be a significant step towards addressing these concerns.

If we have demonstrations of the desired task,
we can extract a reward function using inverse reinforcement learning~\citep{Ng00}.
This reward function can then be used to train an agent with reinforcement learning.
More directly, we can use imitation learning
to clone the demonstrated behavior.
However, these approaches are not directly applicable
to behaviors that are difficult for humans to demonstrate~(such as controlling a robot with many degrees of freedom but very non-human morphology).

An alternative approach is to allow a human to provide feedback
on our system's current behavior and to use this feedback to define the task.
In principle this fits within the paradigm of reinforcement learning,
but using human feedback directly as a reward function
is prohibitively expensive for RL systems that require hundreds or thousands of hours of experience.
In order to practically train deep RL systems with human feedback,
we need to decrease the amount of feedback required by several orders of magnitude.
 
Our approach is to learn a reward function from human feedback
and then to optimize that reward function.
This basic approach has been considered previously,
but we confront the challenges involved in scaling it up to modern deep RL
and demonstrate by far the most complex behaviors yet learned from human feedback.

In summary, we desire a solution to sequential decision problems
without a well-specified reward function that
\begin{enumerate}
\item enables us to solve tasks for which we can only \emph{recognize} the desired behavior,
    but not necessarily demonstrate it,
\item allows agents to be taught by non-expert users,
\item scales to large problems, and
\item is economical with user feedback.
\end{enumerate}

\begin{wrapfigure}{r}{0.5\textwidth}
\begin{center}
\includegraphics[width=0.45\textwidth]{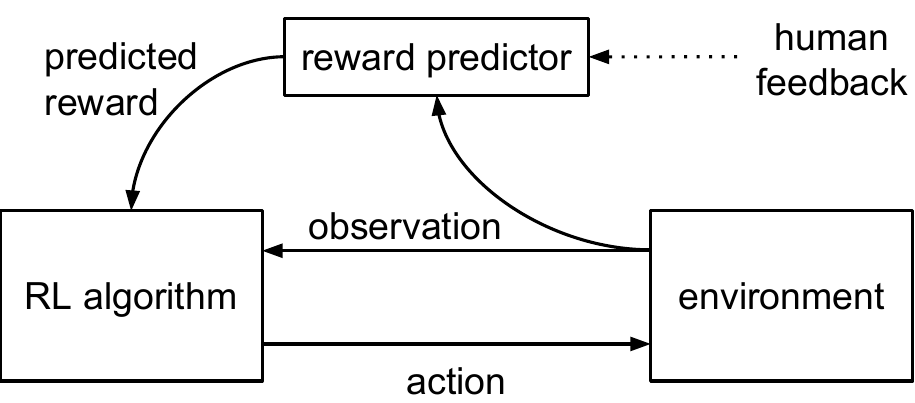}
\end{center}
\caption{Schematic illustration of our approach:
the reward predictor is trained asynchronously from comparisons of trajectory segments,
and the agent maximizes predicted reward.}
\label{fig:reward-feedback-schema}
\end{wrapfigure}

Our algorithm fits a reward function to the human's preferences
while simultaneously training a policy to optimize the current predicted reward function%
~(see \autoref{fig:reward-feedback-schema}).
We ask the human to compare short video clips of the agent's behavior,
rather than to supply an absolute numerical score.
We found comparisons to be easier for humans to provide in some domains,
while being equally useful for learning human preferences.
Comparing short video clips is nearly as fast as comparing individual states,
but we show that the resulting comparisons are significantly more helpful.
Moreover, we show that collecting feedback online improves the system's performance and
prevents it from exploiting weaknesses of the learned reward function.

Our experiments take place in two domains:
Atari games in the Arcade Learning Environment~\citep{Bellemare13},
and robotics tasks in the physics simulator MuJoCo~\citep{Todorov12}.
We show that a small amount of feedback from a non-expert human,
ranging from fifteen minutes to five hours,
suffices to learn most of the original RL tasks even when the reward function is not observable.
We then consider some novel behaviors in each domain,
such as performing a backflip or driving with the flow of traffic.
We show that our algorithm can learn these behaviors from about an hour of feedback---%
even though it is unclear how to hand-engineer a reward function that would incentivize them.

\subsection{Related Work}
\label{ssec:related-work}

A long line of work studies
reinforcement learning from human ratings or rankings,
including \citet{Akrour11}, \citet{Pilarski11}, \citet{Akrour12}, \citet{Wilson12}, \citet{Sugiyama12}, \citet{Wirth13}, \citet{Daniel15}, \citet{ElAsri16}, \citet{Wang16}, and \citet{Wirth16}.
Other lines of research considers the general problem of reinforcement learning
from preferences rather than absolute reward values~\citep{Furnkranz12,Akrour14},
and optimizing using human preferences in settings
other than reinforcement learning~\citep{Machwe06, Secretan08, Brochu10, Sorensen16}.

Our algorithm follows the same basic approach as \cite{Akrour12} and \cite{Akrour14}.
They consider continuous domains with four degrees of freedom and small discrete domains, where they can assume that the reward is linear in the expectations of hand-coded features.
We instead consider physics tasks with dozens of degrees of freedom and Atari tasks with no hand-engineered features;
the complexity of our environments force us to use different RL algorithms and reward models,
and to cope with different algorithmic tradeoffs.
One notable difference is that
\cite{Akrour12} and \cite{Akrour14} elicit preferences over whole trajectories rather than short clips.
So although we gather about two orders of magnitude more comparisons,
our experiments require less than one order of magnitude more human time.
Other differences focus on changing our training procedure
to cope with the nonlinear reward models and modern deep RL,
for example using asynchronous training and ensembling.

Our approach to feedback elicitation closely follows \citet{Wilson12}.
However, \citet{Wilson12} assumes that the reward function
is the distance to some unknown ``target'' policy (which is itself a linear function of hand-coded features).
They fit this reward function using Bayesian inference,
and rather than performing RL they produce trajectories using the MAP estimate
of the target policy.
Their experiments involve ``synthetic'' human feedback which is drawn from their Bayesian model,
while we perform experiments with feedback gathered from non-expert users.
It is not clear if the methods in \citet{Wilson12} can be extended to complex tasks or if they can work with real human feedback.

\citet{MacGlashan17}, \citet{Pilarski11}, \citet{Knox09}, and \citet{Knox12}
perform experiments involving reinforcement learning from actual human feedback,
although their algorithmic approach is less similar.
In \citet{MacGlashan17} and \citet{Pilarski11}, learning only occurs during episodes where the human trainer provides feedback.
This appears to be infeasible in domains like Atari games where thousands of hours of experience
are required to learn a high-quality policy, and would be prohibitively expensive even for the simplest tasks we consider. 
TAMER~\citep{Knox12, Knox13} also learn a reward function,
however they consider much simpler settings where the desired policy can be learned relatively quickly.

Our work could also be seen of a specific instance of the cooperative inverse reinforcement learning framework~\citep{CIRL}.
This framework considers a two-player game between a human and a robot interacting with an environment
with the purpose of maximizing the human's reward function.
In our setting the human is only allowed to interact with this game by stating their preferences.

Compared to all prior work,
our key contribution is to scale human feedback up to deep reinforcement learning and
to learn much more complex behaviors.
This fits into a recent trend of
scaling reward learning methods to large deep learning systems,
for example inverse RL~\citep{Finn16},
imitation learning~\citep{Ho16,Stadie17},
semi-supervised skill generalization~\citep{Finn17},
and bootstrapping RL from demonstrations~\citep{Silver16,Hester17}.

\section{Preliminaries and Method}

\subsection{Setting and Goal}

We consider an agent interacting with an environment over a sequence of steps;
at each time $t$ the agent receives an observation $o_t \in \O$ from the environment
and then sends an action $a_t \in \A$ to the environment.

In traditional reinforcement learning, the environment would also supply a reward $r_t \in \mathbb{R}$
and the agent's goal would be to maximize the discounted sum of rewards.
Instead of assuming that the environment produces a reward signal,
we assume that there is a human overseer who can express preferences between \emph{trajectory segments}.
A trajectory segment is a sequence of observations and actions,
$\sigma = \of{\of{o_0, a_0}, \of{o_{1}, a_{1}}, \ldots, \of{o_{k-1}, a_{k-1}}} \in \of{\O \times \A}^k$.
Write $\sigma^1 \succ \sigma^2$
to indicate that the human preferred trajectory segment $\sigma^1$
to trajectory segment $\sigma^2$.
Informally, the goal of the agent is to produce trajectories which are preferred by the human,
while making as few queries as possible to the human.

More precisely, we will evaluate our algorithms' behavior in two ways:

\begin{description}
\item[Quantitative:] We say that preferences $\succ$ are \emph{generated by} a reward function%
\footnote{Here we assume here that the reward is a function of the observation and action.
In our experiments in Atari environments, we instead assume the reward is a function of the preceding 4 observations.
In a general partially observable environment,
we could instead consider reward functions that depend on the whole sequence of observations,
and model this reward function with a recurrent neural network.}
$r : \O \times \A \rightarrow \mathbb{R}$
if
\[\of{\of{o^1_0, a_0^1},\ldots,\of{o^1_{k-1},a^1_{k-1}}} \succ \of{\of{o^2_{0}, a^2_{0}},\ldots,\of{o^2_{k-1},a^2_{k-1}}}\]
whenever
\[ r\of{o^1_0, a^1_0} + \cdots + r\of{o^1_{k-1}, a^1_{k-1}} > r\of{o^2_{0}, a^2_{0}} + \cdots + r\of{o^2_{k-1}, a^2_{k-1}}.\]
If the human's preferences are generated by a reward function $r$,
then our agent ought to receive a high total reward according to $r$.
So if we know the reward function $r$, we can evaluate the agent quantitatively.
Ideally the agent will achieve reward nearly as high as if it had been using RL to optimize $r$.
\item[Qualitative:] Sometimes we have no reward function by which we can quantitatively evaluate behavior (this is the situation where our approach would be practically useful).
In these cases, all we can do is qualitatively evaluate how well the agent satisfies to the human's preferences.
In this paper, we will start from a goal expressed in natural language,
ask a human to evaluate the agent's behavior based on how well it fulfills that goal,
and then present videos of agents attempting to fulfill that goal.
\end{description}

Our model based on trajectory segment comparisons
is very similar to the trajectory preference queries used in \citet{Wilson12},
except that we don't assume that we can reset the system to an arbitrary state\footnote{\citet{Wilson12} also assumes the ability to sample
reasonable initial states.
But we work with high dimensional state spaces for which random states
will not be reachable and the intended policy inhabits a low-dimensional manifold.}
and so our segments generally begin from different states.
This complicates the interpretation of human comparisons,
but we show that our algorithm overcomes this difficulty even when the human
raters have no understanding of our algorithm.

\subsection{Our Method}

At each point in time our method maintains a policy $\pi : \O \rightarrow \A$
and a reward function estimate $\rest : \O \times \A \rightarrow \mathbb{R}$,
each parametrized by deep neural networks.

These networks are updated by three processes:

\begin{enumerate}
\item The policy $\pi$ interacts with the environment to produce a set of trajectories $\{ \tau^1, \ldots, \tau^i \}$.
    The parameters of $\pi$ are updated by a traditional reinforcement learning algorithm,
    in order to maximize the sum of the predicted rewards $r_t = \rest\of{o_t, a_t}$.
\item We select pairs of segments $\of{\sigma^1, \sigma^2}$
    from the trajectories $\{ \tau^1, \ldots, \tau^i \}$ produced in step 1,
    and send them to a human for comparison.
\item The parameters of the mapping $\rest$ are optimized via supervised learning to fit the comparisons collected from the human so far.
\end{enumerate}

These processes run asynchronously,
with trajectories flowing from process (1) to process (2),
human comparisons flowing from process (2) to process (3),
and parameters for $\rest$ flowing from process (3) to process (1).
The following subsections provide details on each of these processes.

\subsubsection{Optimizing the Policy}
\label{sssec:optimizing-the-policy}

After using $\rest$ to compute rewards, we are left with a traditional reinforcement learning problem.
We can solve this problem using any RL algorithm that is appropriate for the domain.
One subtlety is that the reward function $\rest$ may be non-stationary, which leads us to prefer methods
which are robust to changes in the reward function. 
This led us to focus on policy gradient methods,
which have been applied successfully for such problems~\citep{Ho16}.

In this paper, we use \emph{advantage actor-critic}~(A2C; \citealp{Mnih16}) to play Atari games,
and \emph{trust region policy optimization}~(TRPO; \citealp{Schulman15})
to perform simulated robotics tasks.
In each case, we used parameter settings which have been found to work well for traditional RL tasks.
The only hyperparameter which we adjusted was the entropy bonus for TRPO.
This is because TRPO relies on the trust region to ensure adequate exploration,
which can lead
to inadequate exploration if the reward function is changing.

We normalized the rewards produced by $\rest$ to have zero mean and constant standard deviation.
This is a typical preprocessing step which is particularly appropriate here since the position of the rewards is underdetermined by our learning problem.

\subsubsection{Preference Elicitation}

The human overseer is given a visualization of two trajectory segments,
in the form of short movie clips.
In all of our experiments, these clips are between 1 and 2 seconds long.

The human then indicates which segment they prefer,
that the two segments are equally good, or
that they are unable to compare the two segments.

The human judgments are recorded in a database $\db$ of triples $\of{\sigma^1, \sigma^2, \mu}$,
where $\sigma^1$ and  $\sigma^2$ are the two segments and $\mu$ is a distribution over $\{ 1, 2 \}$ indicating which segment the user preferred.
If the human selects one segment as preferable,
then $\mu$ puts all of its mass on that choice.
If the human marks the segments as equally preferable,
then $\mu$ is uniform.
Finally, if the human marks the segments as incomparable, then the comparison is not included in the database.

\subsubsection{Fitting the Reward Function}
\label{ssec:fitting-reward-function}

We can interpret a reward function estimate $\rest$ as a preference-predictor
if we view $\rest$ as a latent factor explaining the human's judgments
and assume that the human's probability of preferring a segment $\sigma^i$ depends exponentially
on the value of the latent reward
summed over the length of the clip:%
\footnote{\autoref{eq:preference-elo} does not use discounting,
which could be interpreted as modeling the human to be indifferent about when things happen in the trajectory segment.
Using explicit discounting or inferring the human's discount function would also be reasonable choices.}
\begin{equation}\label{eq:preference-elo}
\hat{P}\bdelims*{\sigma^1 \succ \sigma^2} =
\frac{\exp{\sum \rest\of{o^1_t, a^1_t}}}{\exp{\sum\rest\of{o^1_t, a^1_t}} + \exp{\sum\rest\of{o^2_t, a^2_t}}}.
\end{equation}
We choose $\rest$ to minimize the cross-entropy loss
between these predictions and the actual human labels:
\[
  \mathrm{loss}\of{\rest}
= -\sum_{\of{\sigma^1, \sigma^2, \mu} \in \db} \mu\of{1} \log \hat{P}\bdelims*{\sigma^1 \succ \sigma^2}
                                            + \mu\of{2} \log \hat{P}\bdelims*{\sigma^2 \succ \sigma^1}.
\]
This follows the Bradley-Terry model~\citep{Bradley52} for estimating score functions from pairwise preferences,
and is the specialization of the Luce-Shephard choice rule~\citep{Luce05, Shepard57}
to preferences over trajectory segments.
It can be understood as equating rewards with
a preference ranking scale analogous to
the famous Elo ranking system developed for chess~\citep{Elo78}.
Just as the difference in Elo points of two chess players estimates the probability
of one player defeating the other in a game of chess,
the difference in predicted reward of two trajectory segments estimates the probability
that one is chosen over the other by the human.

Our actual algorithm incorporates a number of modifications to this basic approach, which early experiments discovered to be helpful and which
are analyzed in \autoref{ssec:ablation}:
\begin{itemize}
\item We fit an ensemble of predictors, each trained on $\abs{\db}$ triples
    sampled from $\db$ with replacement.
    The estimate $\rest$ is defined by independently normalizing each of these predictors and then averaging the results.
\item A fraction of $1/e$ of the data is held out to be used as a validation set for each predictor.
    We use $\ell_2$ regularization and
    adjust the regularization coefficient to keep the validation loss between $1.1$ and $1.5$ times the training loss.
    In some domains we also apply dropout for regularization.
\item Rather than applying a softmax directly as described in \autoref{eq:preference-elo},
    we assume there is a 10\% chance that the human responds uniformly at random.
Conceptually this adjustment is needed because human raters have a constant probability of making an error, which doesn't decay to 0 as the difference in reward difference becomes extreme.
\end{itemize}

\subsubsection{Selecting Queries}  
\label{sssec:selecting}

We decide how to query preferences based on
an approximation to the uncertainty in the reward function estimator, similar to \citet{Daniel14}:
we sample a large number of pairs of trajectory segments of length $k$,
use each reward predictor in our ensemble
to predict which segment will be preferred from each pair,
and then select those trajectories for which the predictions have the highest variance across ensemble members.
This is a crude approximation and the ablation experiments in \autoref{sec:experiments} show that
in some tasks it actually impairs performance.
Ideally, we would want to query based on the expected value of information of the query~\citep{Akrour12,Krueger16},
but we leave it to future work to explore this direction further.

\section{Experimental Results}
\label{sec:experiments}

We implemented our algorithm in TensorFlow~\citep{Abadi16}.
We interface with MuJoCo~\citep{Todorov12} and the Arcade Learning Environment~\citep{Bellemare13}
through the OpenAI Gym~\citep{Brockman16}.

\subsection{Reinforcement Learning Tasks with Unobserved Rewards}

In our first set of experiments, we attempt to solve a range of benchmark tasks for deep RL \emph{without observing the true reward}.
Instead, the agent learns about the goal of the task only by asking a human which of two trajectory segments is better.
Our goal is to solve the task in a reasonable amount of time using as few queries as possible.

In our experiments, feedback is provided by contractors
who are given a 1-2 sentence description of each task
before being asked to compare several hundred to several thousand pairs of trajectory segments
for that task~(see \autoref{app:instructions} for the exact instructions given to contractors).
Each trajectory segment is between 1 and 2 seconds long.
Contractors responded to the average query in 3-5 seconds,
and so the experiments involving real human feedback
required between 30 minutes and 5 hours of human time.

For comparison, we also run experiments using a synthetic oracle
whose preferences over trajectories exactly reflect reward in the underlying task.
That is, when the agent queries for a comparison,
instead of sending the query to a human, we immediately reply by indicating a preference for
whichever trajectory segment actually receives a higher reward in the underlying task\footnote{In the case of Atari games
with sparse rewards, it is relatively common for two clips to both have zero reward in which case
the oracle outputs indifference.
Because we considered clips rather than individual states,
such ties never made up a large majority of our data.
Moreover, ties still provide significant information to the reward predictor as long as they are not too common.}.
We also compare to the baseline of RL training using the real reward.
Our aim here is not to outperform but rather to do nearly as well as RL
without access to reward information and instead relying on much scarcer feedback.
Nevertheless, note that feedback from real humans does have the potential to outperform RL
(and as shown below it actually does so on some tasks),
because the human feedback might provide a better-shaped reward.

We describe the details of our experiments in \autoref{app:details},
including model architectures,
modifications to the environment,
and the RL algorithms used to optimize the policy.

\subsubsection{Simulated Robotics}

\begin{figure}[t]
\begin{center}
\includegraphics[width=\plotsize\textwidth]{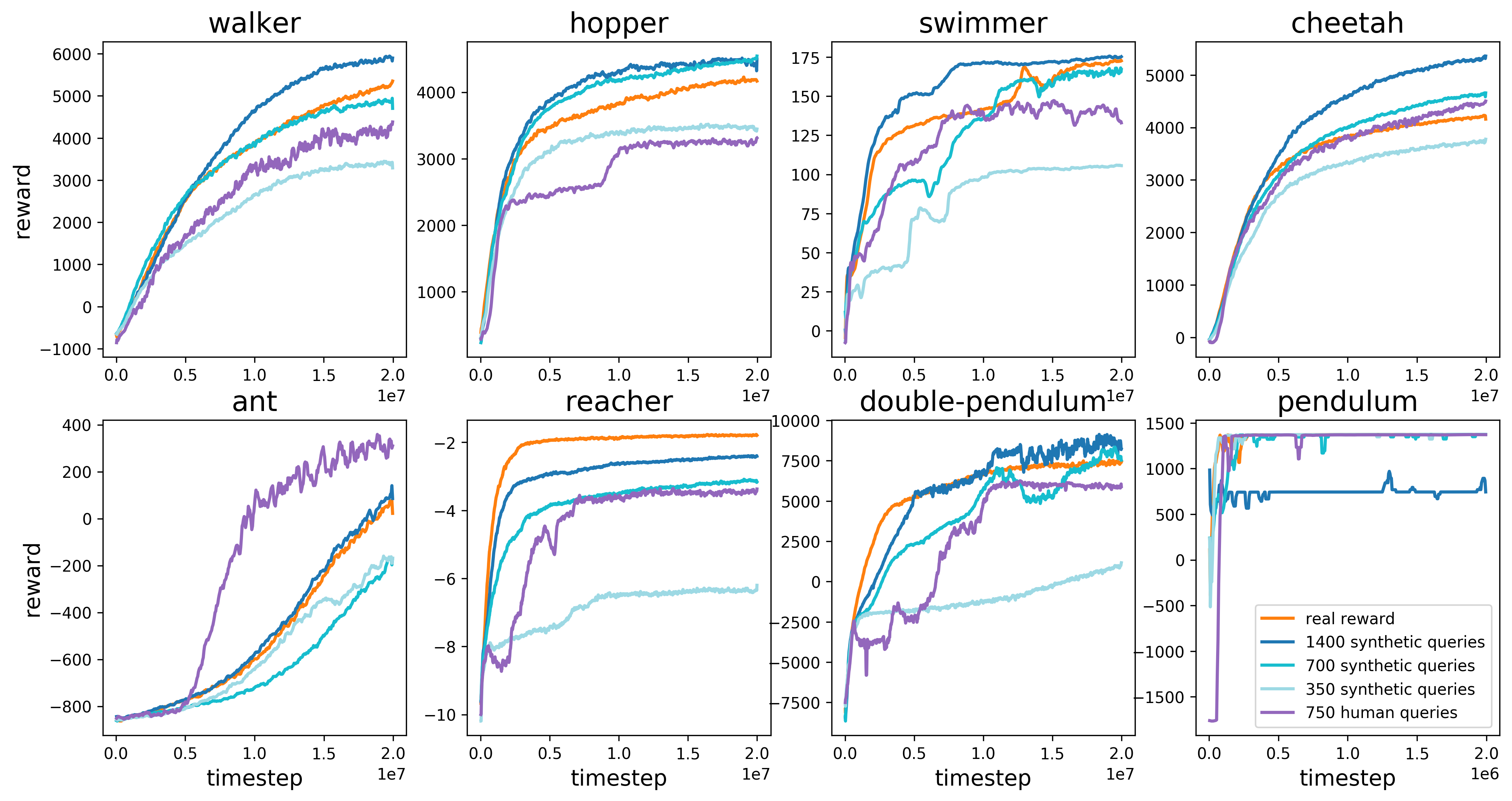}
\end{center}
\caption{%
Results on MuJoCo simulated robotics as measured on the tasks' true reward.
We compare our method using real human feedback~(purple),
our method using synthetic feedback provided by an oracle~(shades of blue), and
reinforcement learning using the true reward function~(orange).
All curves are the average of 5 runs, except for
the real human feedback, which is a single run, and
each point is the average reward over five consecutive batches.
For Reacher and Cheetah
feedback was provided by an author due to time constraints.
For all other tasks,
feedback was provided by contractors
unfamiliar with the environments and with our algorithm.
The irregular progress on Hopper is due to one contractor deviating
from the typical labeling schedule.
}
\label{fig:mujoco-results}
\end{figure}

The first tasks we consider are
eight simulated robotics tasks, implemented in MuJoCo~\citep{Todorov12},
and included in OpenAI Gym~\citep{Brockman16}.
We made small modifications to these tasks
in order to avoid encoding information about the task
in the environment itself (the modifications are described in detail in \autoref{app:details}).
The reward functions in these tasks are linear functions of distances,
positions and velocities, and all are a quadratic function of the features.
We included a simple cartpole task (``pendulum'') for comparison,
since this is representative of the complexity of tasks studied in prior work.

\autoref{fig:mujoco-results} shows the results of training our agent with 700 queries to a human rater,
compared to learning from 350, 700, or 1400 synthetic queries,
as well as to RL learning from the real reward.
With 700 labels we are able to nearly match reinforcement learning on all of these tasks.
Training with learned reward functions tends to be less stable and higher variance,
while having a comparable mean performance.

Surprisingly, by 1400 labels our algorithm performs slightly better than if it had simply been given the true reward,
perhaps because the learned reward function is slightly better shaped---the reward learning procedure assigns positive rewards
to all behaviors that are typically followed by high reward.

Real human feedback is typically only slightly less effective than the synthetic feedback;
depending on the task human feedback ranged from being half as efficient as ground truth
feedback to being equally efficient.
On the Ant task the human feedback significantly outperformed the synthetic feedback, apparently because we asked humans to prefer trajectories where the robot was ``standing upright,'' which proved to be
useful reward shaping. (There was a similar bonus in the RL reward function to encourage the robot to remain upright,
but the simple hand-crafted bonus was not as useful.)

\subsubsection{Atari}

\begin{figure}
\begin{center}
\includegraphics[width=\plotsize\textwidth]{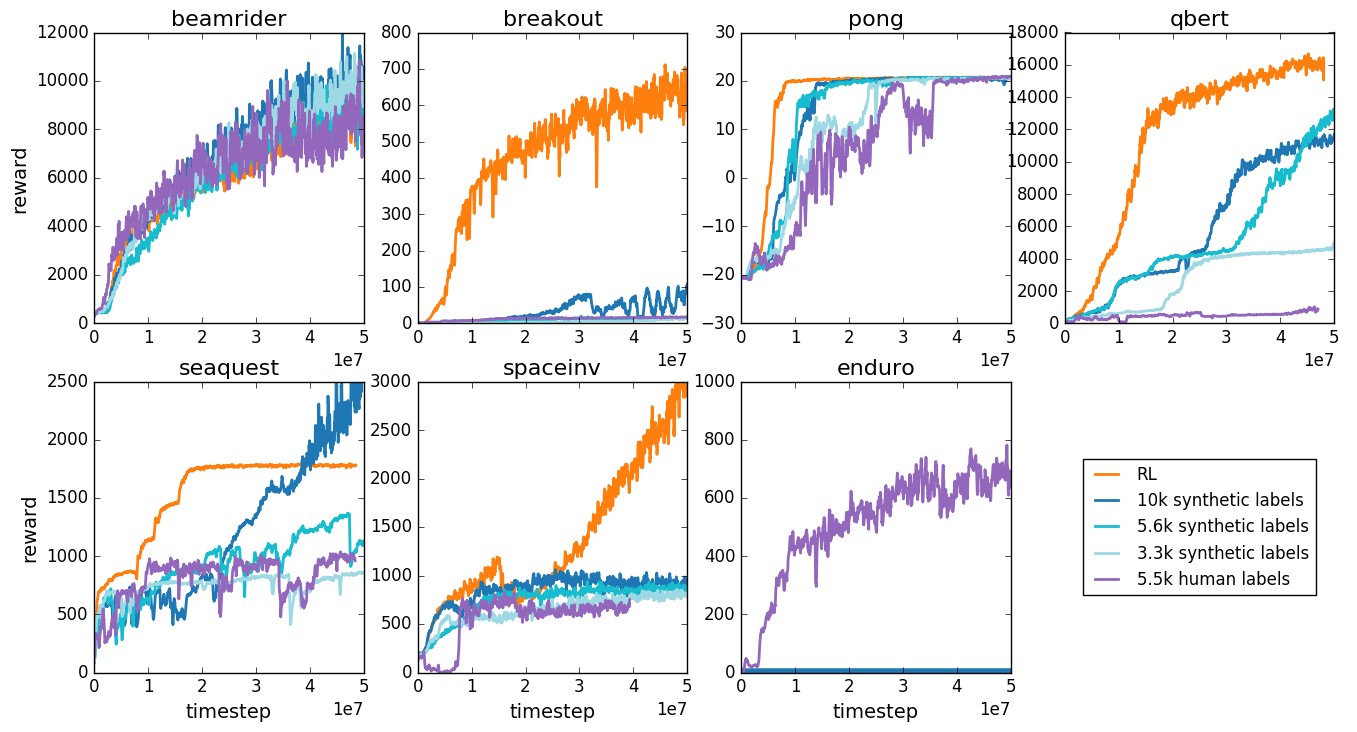}
\end{center}
\caption{%
Results on Atari games as measured on the tasks' true reward.
We compare our method using real human feedback~(purple),
our method using synthetic feedback provided by an oracle~(shades of blue),
and reinforcement learning using the true reward function~(orange).
All curves are the average of 3 runs, except for
the real human feedback which is a single run, and
each point is the average reward over about 150,000 consecutive frames.
}
\label{fig:Atari-results}
\end{figure}

The second set of tasks we consider is a set of
seven Atari games in the Arcade Learning Environment~\citep{Bellemare13},
the same games presented in \citealp{Mnih13}.

\autoref{fig:Atari-results} shows the results of training our agent with 5,500 queries to a human rater,
compared to learning from 350, 700, or 1400 synthetic queries,
as well as to RL learning from the real reward.
Our method has more difficulty matching RL in these challenging environments, but nevertheless it displays substantial learning on most of them and matches or even exceeds RL on some.  Specifically, on BeamRider and Pong, synthetic labels match or come close to RL even with only 3,300 such labels.  On Seaquest and Qbert synthetic feedback eventually performs near the level of RL but learns more slowly.  On SpaceInvaders and Breakout synthetic feedback never matches RL, but nevertheless the agent improves substantially, often passing the first level in SpaceInvaders and reaching a score of 20 on Breakout, or 50 with enough labels.

On most of the games real human feedback performs similar to or slightly worse than synthetic feedback with the same number of labels, and often comparably to synthetic feedback that has 40\% fewer labels.  This may be due to human error in labeling, inconsistency between different contractors labeling the same run, or the uneven rate of labeling by contractors, which can cause labels to be overly concentrated in narrow parts of state space.  The latter problems could potentially be addressed by future improvements to the pipeline for outsourcing labels.  On Qbert, our method fails to learn to beat the first level with real human feedback; this may be because short clips in Qbert can be confusing and difficult to evaluate.  Finally, Enduro is difficult for A3C to learn due to the difficulty of successfully passing other cars through random exploration, and is correspondingly difficult to learn with synthetic labels, but human labelers tend to reward any progress towards passing cars, essentially shaping the reward and thus outperforming A3C in this game~(the results are comparable to those achieved with DQN). 

\subsection{Novel behaviors}\label{sec:novel-behavior}

\newcommand{\link}{\href{https://drive.google.com/drive/folders/0BwcFziBYuA8RM2NTdllSNVNTWTg?resourcekey=0-w4PuSuFvi3odgQXdBDPQ0g&usp=sharing}{this link}}

\begin{figure}[t]
\begin{center}
\includegraphics[width=0.7\textwidth]{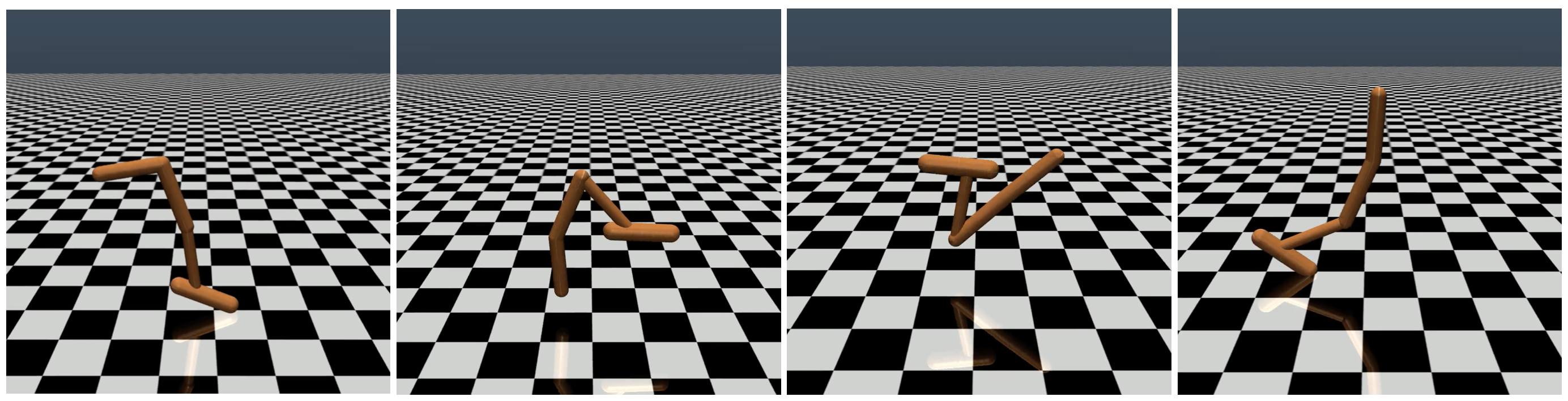}
\end{center}
\caption{Four frames from a single backflip.
The agent is trained to perform a sequence of backflips, landing upright each time.
The video is available \ifanony in the supplementary material\else at \link\fi.}
\label{fig:backflip}
\end{figure}

Experiments with traditional RL tasks help us understand whether our method is effective, but the ultimate purpose of human interaction is to solve tasks for which no reward function is available.

Using the same parameters as in the previous experiments, we show that our algorithm can learn novel complex behaviors. We demonstrate:
\begin{enumerate}
\item The Hopper robot performing a sequence of backflips (see Figure 4). This behavior was trained using 900 queries in less than an hour. The agent learns to consistently perform a backflip, land upright, and repeat.
\item The Half-Cheetah robot moving forward while standing on one leg. This behavior was trained using 800 queries in under an hour.
\item Keeping alongside other cars in Enduro. This was trained with roughly 1,300 queries and 4 million frames of interaction with the environment; the agent learns to stay almost exactly even with other moving cars for a substantial fraction of the episode, although it gets confused by changes in background.
\end{enumerate}
Videos of these behaviors can be found \ifanony in the supplementary material\else  at \link\fi.
These behaviors were trained using feedback from the authors.

\subsection{Ablation Studies}
\label{ssec:ablation}

\begin{figure}[t]
\begin{center}
\includegraphics[width=\plotsize\textwidth]{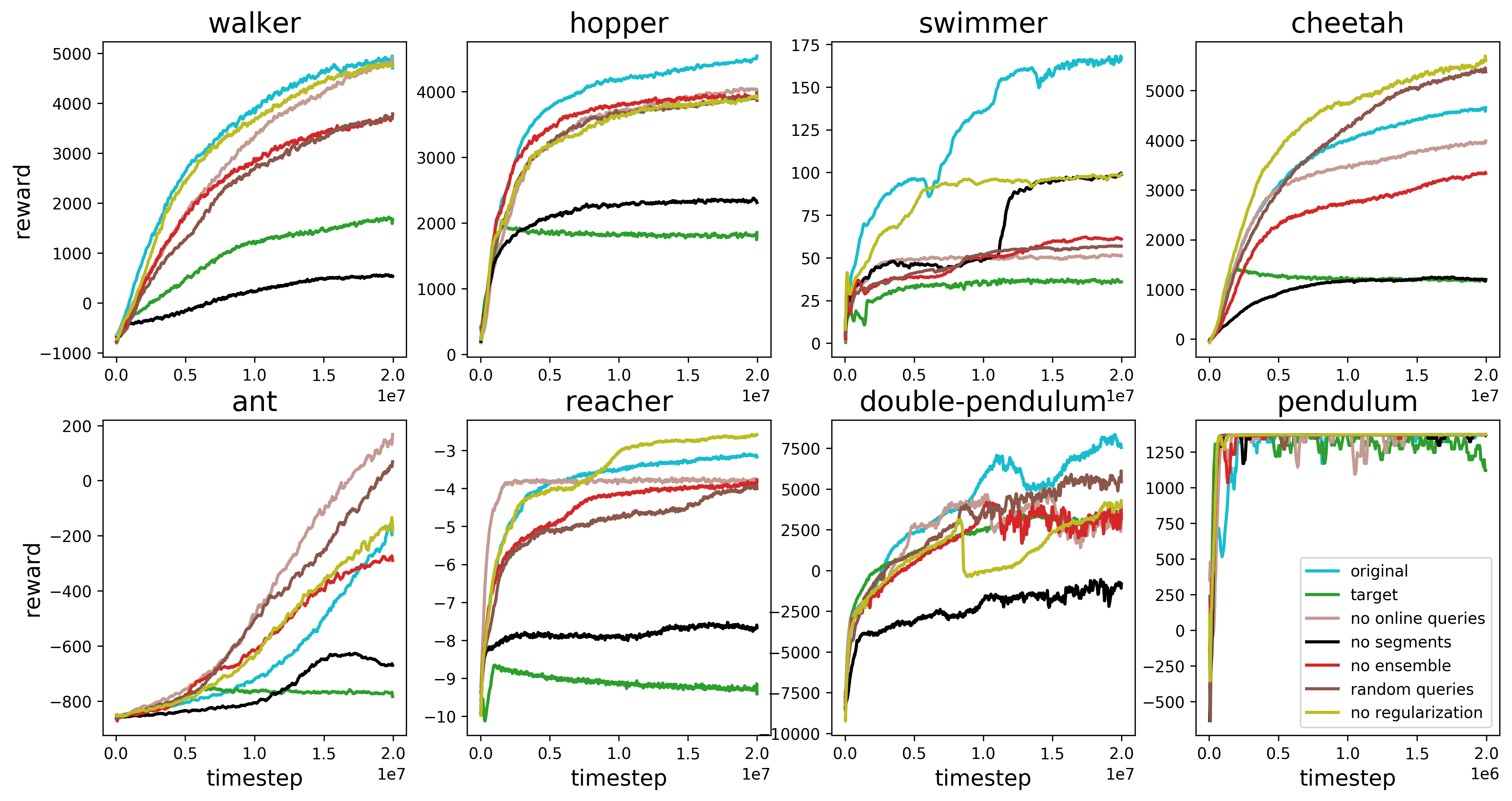}
\end{center}
\caption{Performance of our algorithm on MuJoCo tasks after removing various components, as described in Section~\autoref{ssec:ablation}.
All graphs are averaged over 5 runs, using 700 synthetic labels each. 
}
\label{fig:mujoco_ablation}
\end{figure}

\begin{figure}[t]
\begin{center}
\includegraphics[width=\plotsize\textwidth]{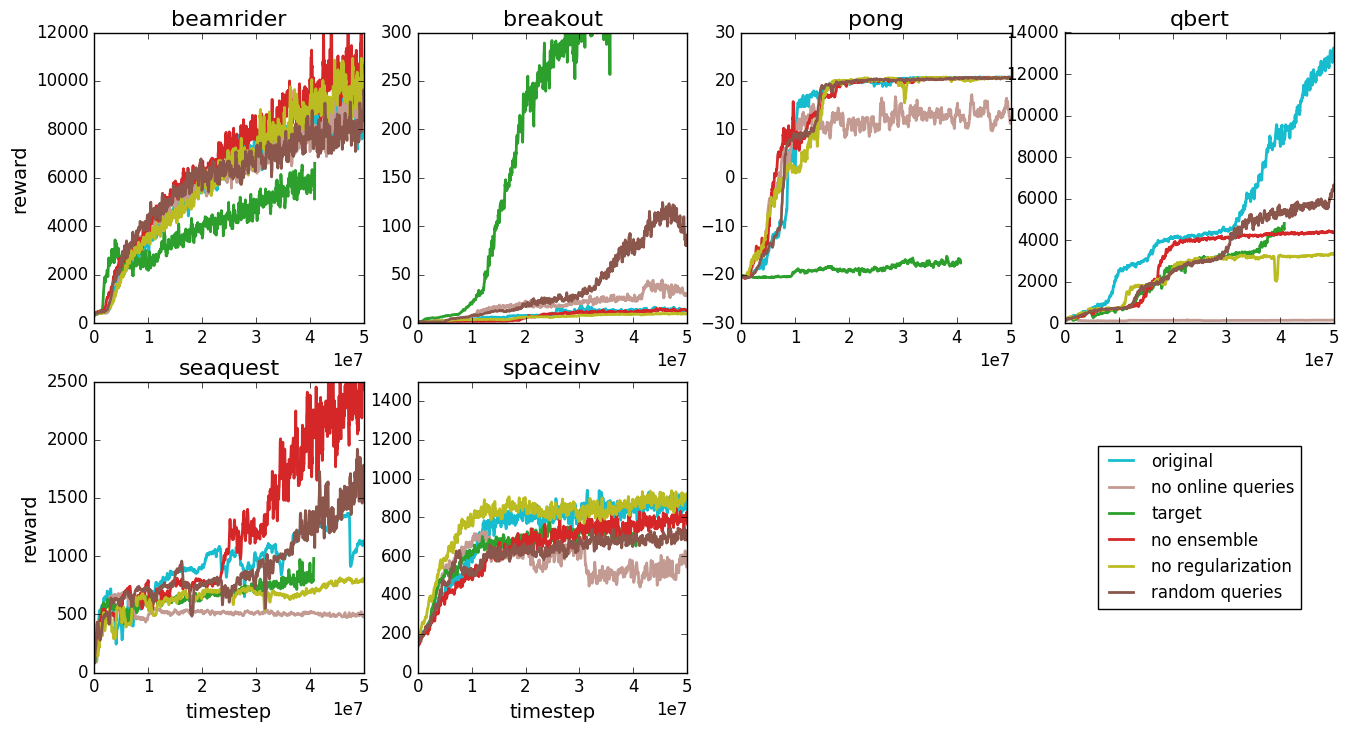}
\end{center}
\caption{Performance of our algorithm on Atari tasks after removing various components, 
as described in \autoref{ssec:ablation}.
All curves are an average of 3 runs using 5,500 synthetic labels (see minor exceptions in \autoref{app:atari-details}).}
\label{fig:Atari-ablation}
\end{figure}

In order to better understand the performance of our algorithm, we consider a range of modifications:
\begin{enumerate}
\item We pick queries uniformly at random rather than prioritizing queries for which there is disagreement~(\textbf{random queries}).
\item We train only one predictor rather than an ensemble~(\textbf{no ensemble}). In this setting, we also choose queries at random, since there is no longer an ensemble that we could use to estimate disagreement.
\item We train on queries only gathered at the beginning of training, rather than gathered throughout training~(\textbf{no online queries}).
\item We remove the $\ell_2$ regularization and use only dropout~(\textbf{no regularization}).
\item On the robotics tasks only, we use trajectory segments of length 1~(\textbf{no segments}).
\item Rather than fitting $\rest$ using comparisons,
we consider an oracle which provides the true total reward over a trajectory segment,
and fit $\rest$ to these total rewards using mean squared error~(\textbf{target}).
\end{enumerate}
The results are presented in
\autoref{fig:mujoco_ablation} for MuJoCo and \autoref{fig:Atari-ablation} for Atari.


Of particular interest is the poor performance of offline reward predictor training;
here we find that due to the nonstationarity of the occupancy distribution,
the predictor captures only part of the true reward,
and maximizing this partial reward can lead to bizarre behavior
that is undesirable as measured by the true reward~\citep{Amodei16}.
For instance, on Pong offline training sometimes leads our agent to avoid losing points but not to score points;
this can result in extremely long volleys that repeat the same sequence of events \emph{ad infinitum}%
~(videos \ifanony in the supplementary material\else at \link\fi).
This type of behavior demonstrates that in general human feedback
needs to be intertwined with RL learning rather than provided statically.

Our main motivation for eliciting comparisons rather than absolute scores was that we found it much easier for humans to provide
consistent comparisons than consistent absolute scores, especially on the continuous control tasks and on the qualitative tasks
in \autoref{sec:novel-behavior}; nevertheless it seems important
to understand how using comparisons affects performance.
For continuous control tasks we found that predicting comparisons worked much better than predicting scores.
This is likely because the scale of rewards varies substantially and this complicates the regression problem,
which is smoothed significantly when we only need to predict comparisons.
In the Atari tasks we clipped rewards and effectively only predicted the sign,
avoiding these difficulties (this is not a suitable solution
for the continuous control tasks because the relative magnitude of the reward are important to learning).
In these tasks comparisons and targets had significantly different performance, but neither consistently outperformed the other.

We also observed large performance differences when using single frames rather than clips\footnote{We only ran these tests on continuous control tasks because our Atari reward model depends on a sequence of consecutive frames rather than a single frame,
as described in \autoref{app:atari-details}}.
In order to obtain the same results using single frames we would need to have collected significantly more comparisons. 
In general we discovered that asking humans to compare longer clips was significantly more helpful \emph{per clip},
and significantly less helpful \emph{per frame}.
We found that for short clips it took human raters a while just to understand the situation,
while for longer clips the evaluation time was a roughly linear function of the clip length.
We tried to choose the shortest clip length for which the evaluation time was linear.
In the Atari environments we also found that it was often easier to compare longer clips because they provide more context
than single frames.

\section{Discussion and Conclusions}

Agent-environment interactions are often radically cheaper than human interaction.
We show that by learning a separate reward model using supervised learning,
it is possible to reduce the interaction complexity by roughly 3 orders of magnitude.
Not only does this show that we can meaningfully train deep RL agents from human preferences,
but also that we are already hitting diminishing returns on further sample-complexity improvements
because the cost of compute is already comparable to the cost of non-expert feedback.%
\footnote{For the Atari experiments we are using a virtual machine with 16~CPUs and one Nvidia~K80 GPU which costs \textasciitilde\textdollar700/month on GCE.
Training takes about a day, so the compute cost is \textasciitilde\textdollar25.
Training with 5k labels corresponds roughly to 5 hours of human labour,
at US minimum wage this totals \textasciitilde\textdollar36.}

Although there is a large literature on preference elicitation and reinforcement learning from unknown reward functions,
we provide the first evidence that these techniques can be economically scaled up to state-of-the-art reinforcement learning systems.
This represents a step towards practical applications of deep RL to complex real-world tasks.

Future work may be able to improve the efficiency of learning from human preferences,
and expand the range of tasks to which it can be applied.

In the long run it would be desirable to make learning a task from human preferences
no more difficult than learning it from a programmatic reward signal,
ensuring that powerful RL systems can be applied in the service of complex human values
rather than low-complexity goals.

\subsubsection*{Acknowledgments}
We thank Olivier Pietquin, Bilal Piot, Laurent Orseau, Pedro Ortega, Victoria Krakovna, Owain Evans, Andrej Karpathy, Igor Mordatch, and Jack Clark for reading drafts of the paper. We thank Tyler Adkisson, Mandy Beri, Jessica Richards, Heather Tran, and other contractors for providing the data used to train our agents.
Finally, we thank OpenAI and DeepMind for providing a supportive research environment and for 
supporting and encouraging this collaboration.

\bibliography{references}

\newpage\appendix
\section{Experimental Details}
\label{app:details}

Many RL environments have termination conditions that depend on the behavior of the agent, such as ending an episode when the agent dies or falls over.
We found that such termination conditions encode information about the task even when the reward function is not observable.
To avoid this subtle source of supervision, which could potentially confound our attempts to learn from human preferences only, we removed all variable-length episodes:
\begin{itemize}
    \item In the Gym versions of our robotics tasks, the episode ends when certain parameters go outside of a prescribed range (for example when the robot falls over). We replaced these termination conditions by a penalty which encourages the parameters to remain in the range (and which the agent must learn).
    \item In Atari games, we do not send life loss or episode end signals to the agent (we do continue to actually reset the environment), effectively converting the environment into a single continuous episode. When providing synthetic oracle feedback we replace episode ends with a penalty in all games except Pong; the agent must learn this penalty.
\end{itemize}
Removing variable length episodes leaves the agent with only the information encoded in the environment itself;
human feedback provides its only guidance about what it ought to do.

At the beginning of training we compare a number of trajectory segments drawn from rollouts of an untrained (randomly initialized) policy.
In the Atari domain we also pretrain the reward predictor for 200 epochs before beginning RL training, to reduce the likelihood of irreversibly learning a bad policy based on an untrained predictor. 
For the rest of training, labels are fed in at a rate decaying inversely with the number of timesteps;
after twice as many timesteps have elapsed,
we answer about half as many queries per unit time.
The details of this schedule are described in each section.
This ``label annealing'' allows us to balance the importance of having a good predictor from the start
with the need to adapt the predictor as the RL agent learns and encounters new states.
When training with real human feedback, we attempt to similarly anneal the label rate,
although in practice this is approximate because contractors give feedback at uneven rates.

Except where otherwise stated we use an ensemble of 3 predictors, and draw a factor 10 more clip pair candidates than we ultimately present to the human, with the presented clips being selected via maximum variance between the different predictors as described in \autoref{sssec:selecting}.

\subsection{Simulated Robotics Tasks}\label{app:mujoco-details}

The OpenAI Gym continuous control tasks penalize large torques.
Because torques are not directly visible to a human supervisor,
these reward functions are not good representatives of human preferences over trajectories and so we removed them.

For the simulated robotics tasks,
we optimize policies using \emph{trust region policy optimization}~(TRPO, \citealp{Schulman15})
with discount rate $\gamma = 0.995$ and $\lambda = 0.97$.
The reward predictor is a two-layer neural network with 64 hidden units each,
using leaky ReLUs~($\alpha = 0.01$) as nonlinearities.\footnote{All of these reward functions are second degree polynomials of the input features, and so if we were concerned only with these tasks we could take a simpler approach to learning the reward function. However, using this more flexible architecture allows us to immediately generalize to tasks for which the reward function is not so simple, as described in \autoref{sec:novel-behavior}.}
We compare trajectory segments that last 1.5 seconds, which varies from 15 to 60 timesteps depending on the task.

We normalize the reward predictions to have standard deviation 1.
When learning from the reward predictor, we add an entropy bonus of 0.01 on all tasks except swimmer, where we use an entropy bonus of 0.001.  As noted in \autoref{sssec:optimizing-the-policy}, this entropy bonus helps to incentivize the increased exploration needed to deal with a changing reward function.

We collect $25\%$ of our comparisons from a randomly initialized policy network at the beginning of training,
and our rate of labeling after $T$ frames
$2*10^6 / (T+2*10^6)$.

\subsection{Atari}\label{app:atari-details}

Our Atari agents are trained
using the standard set of environment wrappers used by \citet{Mnih15}:
0 to 30 no-ops in the beginning of an episode,
max-pooling over adjacent frames, stacking of 4 frames,
a frameskip of 4, 
life loss ending an episode (but not resetting the environment), and
rewards clipped to $[-1, 1]$.

Atari games include a visual display of the score,
which in theory could be used to trivially infer the reward.
Since we want to focus instead on inferring the reward from the complex dynamics happening in the game,
we replace the score area with a constant black background on all seven games.
On BeamRider we additionally blank out the enemy ship count,
and on Enduro we blank out the speedometer. 

For the Atari tasks we optimize policies using the A3C algorithm~\citep{Mnih16} in synchronous form (A2C),
with policy architecture as described in \citet{Mnih15}.
We use standard settings for the hyperparameters:
an entropy bonus of $\beta=0.01$,
learning rate of $0.0007$ decayed linearly to reach zero after 80 million timesteps 
(although runs were actually trained for only 50 million timesteps),
$n=5$ steps per update, $N=16$ parallel workers,
discount rate $\gamma=0.99$,
and policy gradient using Adam with $\alpha=0.99$ and $\epsilon=10^{-5}$. 

For the reward predictor, we use 84x84 images as inputs~(the same as the inputs to the policy),
and stack 4 frames for a total 84x84x4 input tensor.
This input is fed through 4 convolutional layers of size 7x7, 5x5, 3x3, and 3x3 with strides 3, 2, 1, 1,
each having 16 filters, with leaky ReLU nonlinearities~($\alpha = 0.01$).
This is followed by a fully connected layer of size 64 and then a scalar output.
All convolutional layers use batch norm and dropout with $\alpha=0.5$ to prevent predictor overfitting.
In addition we use $\ell_2$ regularization with the adapative scheme described in \autoref{ssec:fitting-reward-function}.
Since the reward predictor is ultimately used to compare two sums over timesteps,
its scale is arbitrary, and we normalize it to have a standard deviation of 0.05~(we could equivalently have adjusted
our learning rates and entropy bonus, but this choice allowed us to use the same parameters as for the real reward function).

We compare trajectory segments of 25 timesteps~(1.7 seconds at 15 fps with frame skipping).

We collect 500 comparisons from a randomly initialized policy network at the beginning of training,
and our rate of labeling after $T$ frames
of training is decreased every $5*10^6$ frames, to be roughly proportional to $5*10^6 / (T+5*10^6)$.

The predictor is trained asynchronously from the RL agent, and on our hardware typically processes 1 label per 10 RL timesteps.  We maintain a buffer of only the last 3,000 labels and loop over this buffer continuously; this is to ensure that the predictor gives enough weight to new labels (which can represent a shift in distribution) when the total number of labels becomes large.

In the ablation studies of Figure 5b, pretraining has 5,000 labels rather than 5,500, 
and the ``target'' beamrider curve is averaged over 2 runs rather than 3.

\section{Instructions Provided to Contractors}
\label{app:instructions}

\subsection{MuJoCo}

\subsubsection*{Giving feedback}

Sign up for a slot in the spreadsheet.
Then go to the appropriate URL's that we give you,
and you'll be repeatedly presented with two video clips of an AI controlling a virtual robot.

\textbf{Look at the clips and select the one in which better things happen.}
Only decide on events you actually witness in the clip.

\textbf{Here’s a guide on what constitutes good and bad behavior in each specific domain:}
\begin{itemize}
\item \textbf{Hopper}: the ``center'' of the robot is the joint closest to the pointy end.
The first priority is for the center of the robot to move to the right~(moving to the left is worse than not moving at all).
If the two robots are roughly tied on this metric, then the tiebreaker is how high the center is.
\item \textbf{Walker}: the ``center'' of the robot is the joint where the three limbs meet.
The first priority is for the center of the robot to move to the right.
If the two robots are roughly tied on this metric, then the tiebreaker is how high the center is.
\item \textbf{Swimmer}: the ``center'' of the robot is the mark in the middle of its body.
The center should move to the right as fast as possible.
\item \textbf{Cheetah}: the robot should move to the right as fast as possible.
\item \textbf{Ant}: the first priority is for the robot to be standing upright,
and failing that for the center of the robot to be as high up as possible.
If both robots are upright or neither is, the tie breaker is whichever one is moving faster to the right.
\item \textbf{Reacher}: the green dot on the robot arm should be as close as possible to the red dot.
Being near for a while and far for a while is worse than being at an intermediate distance for the entire clip.
\item \textbf{Pendulum}: the pendulum should be pointing approximately up.
There will be a lot of ties where the pendulum has fallen and a lot of ``can't tells'' where it is off the side of the screen.
If you can see one pendulum and it hasn't fallen down,
that’s better than being unable to see the other pendulum.
\item \textbf{Double-pendulum}: both pendulums should be pointing approximately up~(if they fall down,
the cart should try to swing them back up)
and the cart should be near the center of the track.
Being high for a while and low for a while is worse than being at an intermediate distance the entire time.
\end{itemize}

If both clips look about the same to you, then click ``tie''.
If you don't understand what's going on in the clip or find it hard to evaluate, then click ``can't tell''.

\textbf{You can speed up your feedback by using the arrow keys} \\
\texttt{left} and \texttt{right} select clips, \texttt{up} is a tie, \texttt{down} is ``can't tell''.

\subsubsection*{FAQ}

\textbf{I got an error saying that we’re out of clips. What’s up?}
Occasionally the server may run out of clips to give you, and you'll see an error message.
This is normal, just wait a minute and refresh the page.
If you don't get clips for more than a couple minutes, please ping @tom on slack.

\textbf{Do I need to start right at the time listed in the spreadsheet?}
Starting 10 minutes before or after the listed time is fine.

\subsection{Atari}

\begin{quote}\it
In this task you’ll be trying to teach an AI to play Atari games
by giving it feedback on how well it is playing.
\end{quote}

\subsubsection*{IMPORTANT. First play the game yourself for 5 minutes}

Before providing feedback to the AI,
play the game yourself for a five minutes to get a sense of how it works.
It's often hard to tell what the game is about just by looking at short clips,
especially if you've never played it before.

Play the game online for 5 minutes.\footnote{e.g. \url{http://www.free80sarcade.com/2600_Beamrider.php}}
You’ll need to press \texttt{F12} or click the \texttt{GAME RESET} button to start the game.
Then set a timer for 5 minutes and explore the game to see how it works.

\subsubsection*{Giving feedback}

Sign up for a slot in the spreadsheet.
Then go to the appropriate URL's that we give you,
and you'll be repeatedly presented with two video clips of an AI playing the game.

\textbf{Look at the clips and select the one in which better things happen.}
For example, if the left clip shows the AI shooting an enemy ship
while the right clip shows it being shot by an enemy ship,
then better things happen in the left clip and thus the left clip is better.
Only decide on actions you actually witness in the clip.

\textbf{Here’s a guide on what constitutes good and bad play in each specific game:}

\begin{itemize}
\item \textbf{BeamRider}: shoot enemy ships (good), and don’t get shot (very bad)
\item \textbf{Breakout}: hit the ball with the paddle, break the colored blocks, and don’t let the ball fall off the bottom of the screen
\item \textbf{Enduro}: pass as many cars as you can, and don’t get passed by cars
\item \textbf{Pong}: knock the ball past the opponent’s orange paddle on the left (good), and don’t let it go past your green paddle on the right (bad)
\item \textbf{Qbert}: change the color of as many blocks as you can (good), but don’t jump off the side or run into enemies (very bad)
\item \textbf{SpaceInvaders}: shoot enemy ships (good), and don’t let your ship (the one at the bottom of the screen) get shot (very bad)
\item \textbf{SeaQuest}: Shoot the fish and enemy submarines (good) and pick up the scuba divers. Don’t let your submarine run out of air or get hit by a fish or torpedo (very bad)
\item \textbf{Enduro (even mode)}: Avoid passing cars OR getting passed by them, you want to stay even with other cars (not having any around is OK too)
\end{itemize}

\textbf{Don't worry about how the agent got into the situation it is in}
(for instance, it doesn't matter if one agent has more lives,
or is now on a more advanced level);
just focus on what happens in the clip itself. 

If both clips look about the same to you, then click ``tie''.
If you don't understand what's going on in the clip or find it hard to evaluate,
then click ``can't tell''.
Try to minimize responding ``can't tell'' unless you truly are confused.

\textbf{You can speed up your feedback by using the arrow keys} \\
\texttt{left} and \texttt{right} select clips, \texttt{up} is a tie, \texttt{down} is ``can't tell''.

\subsubsection*{FAQ}

\textbf{I got an error saying that we’re out of clips. What’s up?}
Occasionally the server may run out of clips to give you, and you'll see an error message.
This is normal, just wait a minute and refresh the page.
If you don't get clips for more than a couple minutes, please ping @tom on slack.

\textbf{If the agent is already dead when the clip starts, how should I compare it?}
If the clip is after getting killed (but not showing the dying),
then its performance during the clip is neither good nor bad.
You can treat it as purely average play.
If you see it die, or it’s possible that it contains a frame of it dying, then it’s definitely bad.

\textbf{Do I need to start right at the time listed in the spreadsheet?}
Starting 30 minutes before or after the listed time is fine.

\end{document}